# ML-based Predictive Models for Power Consumption in Virtualised O-RANs

Rishu Raj, Genevieve Akude, Urooj Tariq, and Daniel Kilper
CONNECT Centre, Trinity College Dublin, Ireland

*Abstract* — **As communication networks adopt virtualized and disaggregated architectures, achieving energy efficiency has become increasingly important for both economic and environmental reasons. Traditional methods for power modeling are inadequate in these dynamic software-defined environments due to their inability to model complex and nonlinear factors affecting energy use. We investigate the use of feature extraction and regressor-based machine learning methods for predicting power consumption in virtualized open radio access networks (O-RANs), utilizing datasets from a hardware-instrumented testbed. We test three variants of deep neural networks (DNNs), namely, a standard DNN, a regularized DNN, and a hybrid model combining DNN-based feature extraction with an XGBoost regressor. We evaluate the performance of these models for various system parameters such as transmission gain, modulation/coding schemes, and airtime. We show that the hybrid model consistently outperformed others, achieving a mean relative error below 0.5%. Results suggest hybrid models like DNN-XGBoost offer superior accuracy and could be integrated into O-RAN management tools to enable more energy-efficient network orchestration in future networks.**

## I. INTRODUCTION

The energy consumption of mobile networks has become one of the most pressing challenges for operators, particularly with the large-scale rollout of fifth-generation (5G) and beyond communication networks. As communication infrastructures increase in both scale and functional complexity, energy demands contribute to substantial greenhouse gas emissions and significantly higher operational expenditure [1]. The introduction of 5G has aggravated this challenge due to network densification (driven by the deployment of additional small cells) and the inclusion of radios operating in newly allocated, higher-frequency bands such as millimetre wave. Consequently, there has been a marked rise in power requirements, particularly in the radio access networks (RANs), which account for over 70% of the total energy consumption in mobile systems [2]. A study that analysed 31 global operators showed that the radio units (RUs) are alone responsible for ~70% of the power consumed in a site. Within RUs, the power amplifier is the dominant consumer, accounting for 59% of RU energy usage [2]. These statistics underline the importance of targeted optimisation strategies for specific RAN components, as small improvements in these areas can yield substantial overall energy savings.

Open radio access network (O-RAN) architectures, as defined by the O-RAN Alliance [1], provide an opportunity to address this challenge through openness, modularity, and disaggregation. O-RAN facilitates the integration of multi-vendor components and enables more flexible network operation. This architectural shift supports dynamic reconfiguration of network elements based on traffic patterns and load conditions, which can be leveraged to improve energy efficiency in line with global sustainability goals such as the EU's 2050 Net-Zero target. However, the highly virtualised and reconfigurable nature of O-RAN also introduces additional complexity in modelling and predicting energy consumption due to the complexity of coordinating power requirements across multiple, semi-independent components. Each node's energy usage varies in response to fluctuating network conditions. For example, RUs require higher power during peak hours to accommodate traffic surges, but lower power during off-peak periods. To this end, researchers have proposed analytical models to predict the power consumption in O-RAN.

The EARTH model [3] is a component-level power consumption model, which was initially developed for traditional RANs, where it estimated the energy consumption of base stations. In [4], the authors adapted this model for O-RAN nodes to accommodate the disaggregated nature of O-RAN architectures. They also proposed another model for power consumption in RUs that employ carrier aggregation for throughput enhancement [5]. Another model proposed in [6] categorises specific operational states and uses the time spent in each state to compute overall energy consumption. However, analytical models are insufficient for O-RAN power prediction because these models have a linear structure with hardware-agnostic parameters that fail to capture the practical features of O-RAN, like disaggregation, functional-split choices, accelerator heterogeneity, near-real-time control, and discrete energy states [7]. Power consumption in O-RAN is driven by discrete energy states, real-time scheduler bursts, and platform-specific acceleration modes rather than smooth load–power relationships [3]. Recent deployments of virtual radio access network (vRAN) show server-side timing constraints invalidate generic central processing unit (CPU) power-governor assumptions, shifting the load–power relationship beyond classical models [1].

Machine learning (ML), particularly deep neural networks (DNNs), offers a promising solution to the limitations of analytical models in predicting power consumption in O-RAN systems. DNN-based models can learn intricate, data-driven relationships from historical telemetry, enabling accurate prediction under varying operational states, scheduler behavior, and platform-specific energy modes. This data-centric adaptability is crucial for capturing discontinuities in power consumption caused by bursty traffic, real-time constraints, and deployment-specific configurations, where classical models fail. The use of DNN architectures to predict the performance and energy efficiency has garnered significant attention in the research on O-RAN. Authors in [8] demonstrated the ability of neural networks to configure performance mappings and align them for energy modelling using Bayesian learning to optimise power performance trade-offs in virtualized base stations (vBSs). In [9], researchers developed a black-box neural network to predict vBS energy use, highlighting the flexibility of DNNs in modelling complex systems like O-RAN. The work presented in [10] explores DNN-based short-term energy forecasting models, emphasising how hybrid models (where DNNs are combined with signal decomposition techniques or external controllers) improve accuracy and stability. Such hybrid architectures are relevant in O-RAN where energy trends can vary with parameter configurations. The authors in [11] conclude that hybrid models which combine DNN with techniques such as XGBoost achieve optimal results, especially for highly variable environments. Researchers in [12] develop a deep energy predictor model, a hybrid approach combining XGBoost and DNN. The model leveraged the strengths of both techniques to enhance the efficiency of electrical consumption predictions in a household.

In this work, we provide empirical insights into the relationship between system configurations and energy usage by leveraging the dataset [13] collected on an O-RAN testbed by Salvat *et al.* in [14]. We design and implement deep learning models to predict the power consumption in a virtualised O-RAN testbed: (1) a baseline DNN model, (2) a regularised DNN model, and (3) a hybrid DNN–XGBoost pipeline, where the DNN serves as a feature extractor and XGBoost performs the final prediction. We apply advanced feature engineering techniques to improve the input quality and predictive performance of these models. We evaluate model robustness and generalisability by analysing grouped subsets of system configurations. We study the predictive performance of these models across different system configurations, including transmission gain levels, modulation and coding schemes, and airtime allocations. This work aims to establish a benchmark for AI-based energy prediction in O-RAN systems and provide actionable insights for intelligent energy management in next-generation mobile networks.

## II. Virtualised O-RAN System

In this section, we present the architecture of the virtualised O-RAN system, including its functional components, infrastructure, and power measurement methodology as described in [14]. The dataset utilised in this project is sourced from the publicly available experimental evaluation platform [13], collected in controlled testbed environments to ensure reproducibility, with metrics logged directly from the software's physical (PHY) and medium access control layers.

The published configuration, shown in Fig. 1, follows O-RAN principles by disaggregating the base station into virtualised components on a shared compute platform, enabling fine-grained monitoring and flexible experimentation with radio and compute configurations. Implemented on a commodity x86 Ubuntu server with Docker containerisation, it supports uplink and downlink traffic via laptops acting as user equipment (UEs). In accordance with O-RAN specifications, the RU is realised with a universal software radio peripheral (USRP) B210 software-defined radio, and handles lower PHY operations (signal processing, amplification, fast Fourier transform) through the USRP hardware driver and an open-source LTE/5G PHY stack. The DU and CU are deployed as virtualised network functions on an Intel i7 O-RAN cloud platform using Docker for isolation and scalability, with vBS instances pinned to dedicated central CPU cores to emulate realistic scheduling. The platform supports real-time telemetry of CPU utilisation and resource scheduling, enabling experiments under controlled load variations and accurate compute-energy measurement [14].

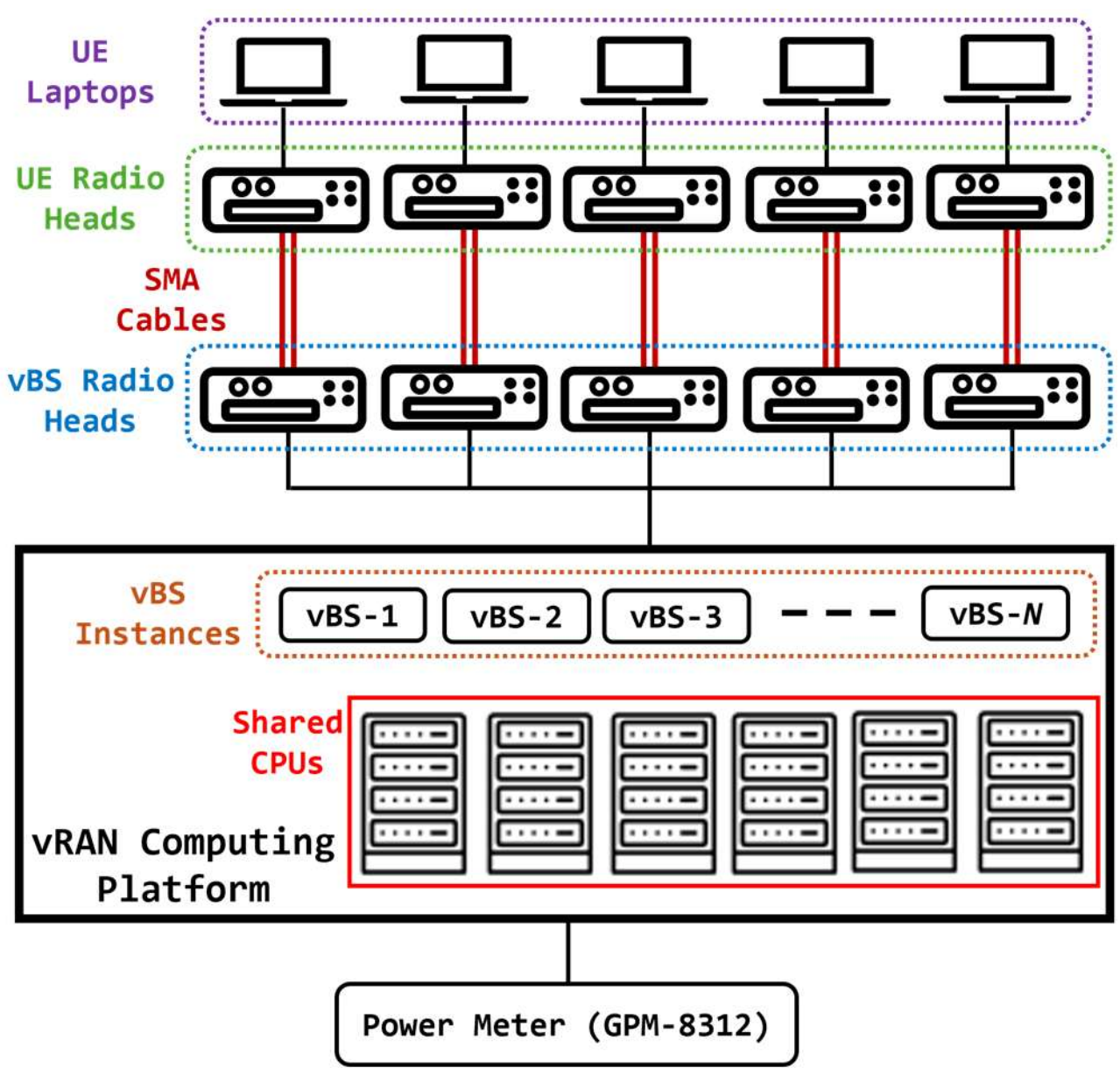

Fig. 1. Virtualized O-RAN model used in [13] for power consumption measurements.

The GW-Instek GPM-8213 power meter, connected to the server's power supply, measures system-wide (RU, CPU, RAM, and background services) power consumption via the standard commands for programmable instruments protocol [14]. The measurement workflow begins with an edge server generating controlled UL/DL traffic to UEs, transmitted via the RU, with transmission gain and modulation parameters logged. Signals are processed by vBS containing DU/CU instances sharing physical CPU cores. Despite the structured collection process, measurements were affected by system-level noise from CPU scheduling variability, thermal effects, and memory access latency. Identical configurations yielded slight power differences.

## III. DNN-based Predictive Models

In this section, we present our implementation of the DNN models under study for predicting power consumption in a virtualised O-RAN environment. The implementation comprises systematic data preprocessing followed by the iterative evolution of the DNN architecture.

### *A. Data Preprocessing*

The first stage of model implementation involves extensive data preprocessing to ensure that the data is suitable for training and testing in the DNNs. Preprocessing of data encompasses feature engineering aimed at extracting and refining informative attributes, and dataset partitioning, which ensures proper separation into training and testing subsets, thereby supporting effective model development and evaluation.

#### *1) Feature Engineering*

We perform feature engineering to extract high-value features that capture the complex, nonlinear relationships between system configuration and power consumption.

Feature engineering involves transforming and selecting variables that have a strong influence on the target prediction. The target variable is average system power consumption in watts. To identify key predictors, we apply permutation feature importance [15], which measures the increase in mean absolute error when feature values are shuffled. The analysis shows transmission gain, modulation/coding schemes, and airtime as the strongest predictors, while features such as user equipment count and clock speed add limited value. This analysis informed the final feature selection, helping reduce model complexity while preserving predictive performance.

*2) Dataset Partitioning*

We design a partitioning strategy to ensure fair training and testing while analysing prediction performance under varied network conditions. Instead of a single random split across the entire dataset for training and testing, we first group the full dataset according to configuration parameters like transmission gain, airtime, modulation scheme etc. After grouping the data into homogeneous subsets based on these parameters, each group was independently partitioned into training (80%) and testing (20%) subsets. Within each group, we split the training set further, reserving 10% for validation. This maintains the reliability of the evaluation process as it prevents data leakage, which is particularly critical when working with measured data from controlled testbed environments that have noise.

### *B. Evolution of DNN Architecture*

To model the complex and nonlinear relationships between O-RAN configuration parameters and system-level power consumption, we iteratively develop a series of DNN architectures. Each successive model incorporates insights from its predecessor, addresses observed shortcomings, and accounts for the nonlinear characteristics of the data. This section presents the evolution of the modeling approach in three stages. Each model trains on the configuration-specific dataset partitions described in the previous sub-section. We track performance across the validation and test data subsets.

*1) Model 1: Baseline DNN Model*

We first design a simple feedforward neural network that serves as a performance baseline. The objective is to construct a foundational architecture that captures the essential nonlinear dependencies between input features and power consumption, without introducing advanced regularization or optimization heuristics. The architecture consists of an input layer with neurons equal to the number of engineered input features, followed by three fully connected hidden layers with 64, 64, and 32 neurons, respectively. The two 64-neuron layers allow exploration of a wide range of representations while retaining useful features, whereas the 32-neuron layer reduces dimensionality to eliminate redundancy while preserving essential patterns. We apply the rectified linear unit (ReLU) activation to mitigate vanishing gradients and ensure stable gradient updates, and employ the Adam optimizer to balance convergence speed and stability by adaptively tuning learning rates [16]. The loss function is the mean squared error (MSE), and the output layer consists of a single neuron with linear activation to predict continuous power consumption. We use a batch size of 64, chosen to balance gradient stability against generalization. We train the model on standardized input features. The baseline achieves promising training performance but overfits, which is expected due to the absence of weight constraints.

*2) Model 2: Regularised DNN Model*

To address the overfitting observed in the baseline model, we design a second architecture with structural and optimization-level regularization mechanisms. We increase the number of neurons in the first hidden layer to 128 to capture more complex patterns, while the other two layers have the same number of neurons as in Model 1. We apply dropout after each hidden layer, which randomly deactivates neurons with probability $p = 0.3$. The dropout operation [17] is defined as $h(l) = r(l) \odot \tilde{h}(l)$, where $\tilde{h}(l)$ is the pre-dropout activation of the layer $l$, $\odot$ denotes element-wise multiplication, and $r(l)$ is the random dropout mask, such that $r(l) \sim \text{Bernoulli}(1-p)$. We also add L2 regularization (weight decay) [18] to the loss function with coefficient $\lambda = 0.01$. The regularized loss is given as

$$\mathcal{L}_{\text{reg}} = \mathcal{L}_{\text{MSE}} + \lambda \sum_j w_j^2 \quad (1)$$

where $w_j$ denotes the model parameters. Batch normalization is applied before each ReLU activation to normalize activations across mini-batches and reduce internal covariate shift. These design choices enhance model stability across grouped configurations. Regularization flattens the loss landscape, leading to better validation accuracy, reduced variance, and slower overfitting.

*3) Model 3: Hybrid DNN–XGBoost Model*

Although the regularized model reduces overfitting and stabilizes training, it fails to capture the full complexity of power consumption across diverse system configurations. Neural networks effectively model smooth nonlinear dependencies but struggle to represent rule-based behaviors, such as abrupt power surges under specific modulation levels or airtime saturation. Moreover, the black-box nature of neural models limits interpretability, which is essential in real-world mobile network deployments. To overcome these limitations, we design a hybrid model that combines the representation learning capacity of a DNN with the structured decision-making ability of XGBoost [11]. This two-stage model separates representation and decision phases. The DNN serves as a high-capacity feature extractor, mapping raw input parameters into a lower-dimensional latent space. These embeddings are then passed to an XGBoost regressor [19], which maps them to accurate power consumption estimates.

The DNN architecture comprises four dense layers with 587, 261, 186, and 99 units, respectively. Each layer employs L2 regularization with coefficient $\lambda = 0.01$. A bottleneck layer with 16 neurons produces compact embeddings, which form the input to the XGBoost model. We configure XGBoost with a maximum depth of $d = 5$, number of estimators $T = 256$, and learning rate $\eta = 0.22$. These values have been selected empirically to balance complexity

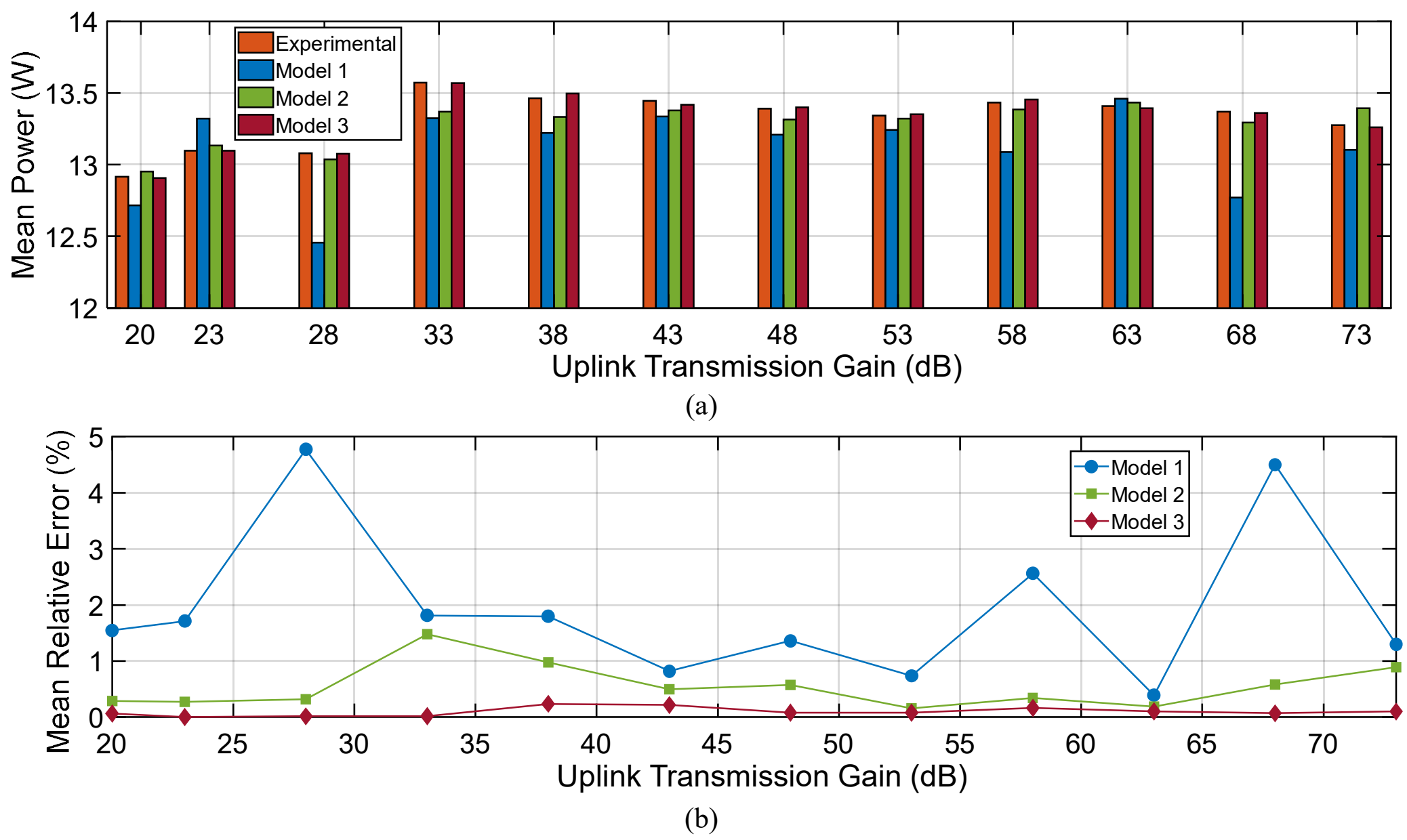

Fig. 2. Variation in (a) power consumption, and (b) relative error with increase in uplink transmission gain for different predictive models.

with generalization. XGBoost minimizes the following regularized objective function [19]:

$$\mathcal{L}_{\mathrm{XGB}} = \sum_{i=1}^{N} l\left(y_i, \hat{y}_i^{(t)}\right) + \sum_{t=1}^{T} \Omega(f_t) \tag{2}$$

where $l(\cdot)$is the differentiable loss function, $y_i$ is the actual value, $\hat{y}_i^{(t)}$ is the prediction from tree $t$, $N$ is the number of training samples, and $\Omega(f_t)$ is the regularization term that penalizes model complexity. The training pipeline proceeds in two phases. First, the DNN trains on standardized features with MSE loss until convergence. We then freeze its parameters and extract the 16-dimensional embeddings from the bottleneck layer. In the second phase, these embeddings are used as input to the XGBoost regressor. This modular approach enables the DNN to specialize in feature abstraction while XGBoost focuses on high-fidelity prediction via rule-based logic and fine-grained splits.

## IV. Results and Discussion

In this section, we present a comparative evaluation of the three predictive models proposed in this work, namely, baseline DNN (Model 1), regularized DNN (Model 2), and a hybrid DNN–XGBoost pipeline (Model 3). The discussion is organized by configuration parameters that strongly affect power consumption in the virtualised O-RAN testbed. These are transmission gain, modulation and coding scheme (MCS) index, and airtime. In each case, we report the mean relative percentage error (MRE) which is interpreted with respect to expected system behavior. The MRE for the $j^{th}$ configuration parameter is computed as

$$\varepsilon_j = \frac{1}{n_j} \sum_{i=1}^{n_j} \left| \frac{x_{i,j} - \hat{x}_{i,j}}{x_{i,j}} \right| \times 100\% \tag{3}$$

where $n_j$ is the number of experiments, $x_i$ and $\hat{x}_i$ are the experimental and predicted values, respectively, of the $i^{th}$ experiment.

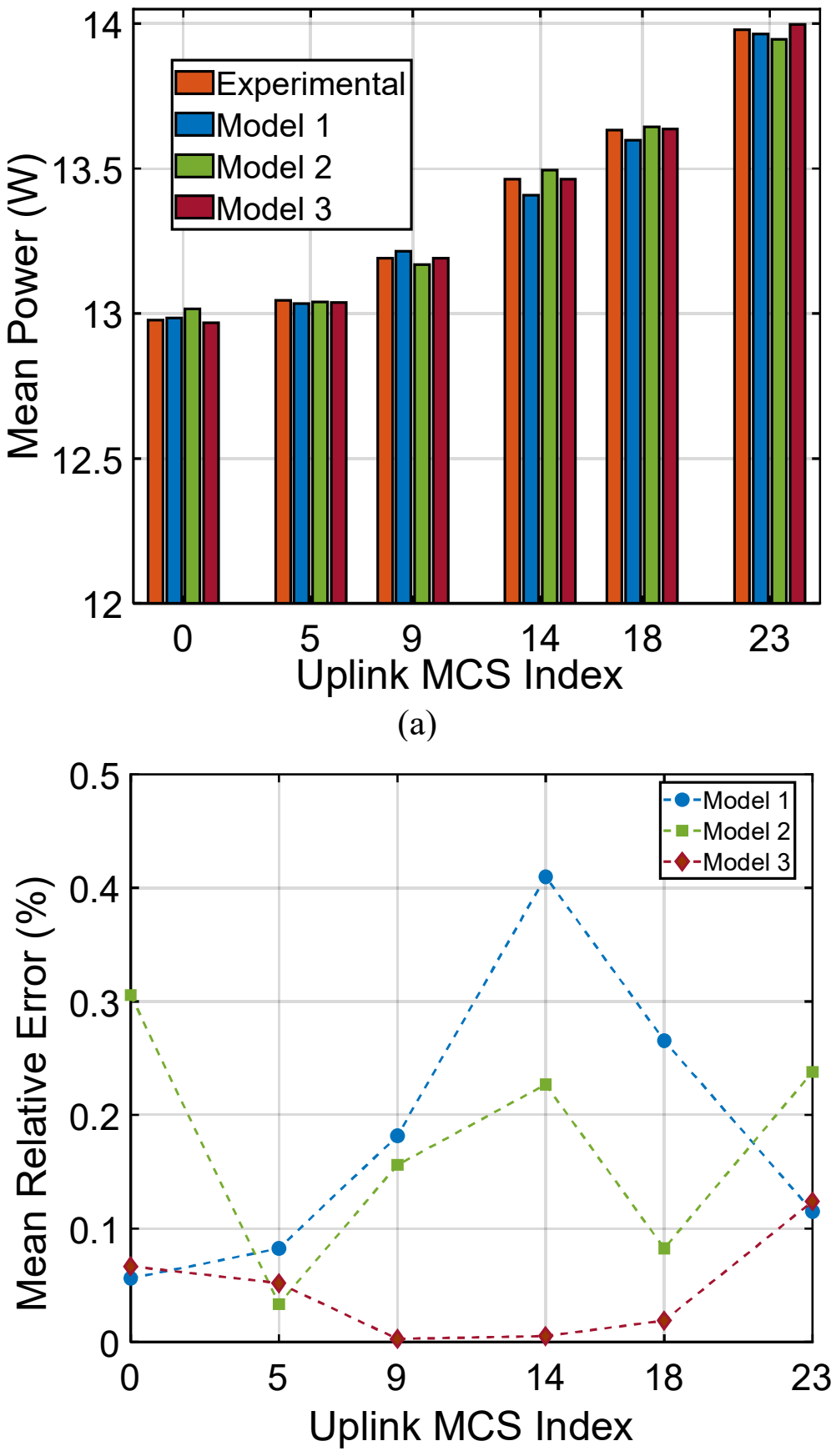

Fig. 3. Variation in (a) power consumption, and (b) relative error with increase in uplink modulation and coding scheme (MCS) index for different predictive models.

### A. Uplink Transmission Gain

In Fig. 2(a), we analyse the power consumption predicted by the three values at various gain settings between 20 dB and 73 dB. As shown in Fig. 2(b), Model 1 (baseline DNN) produced MRE values between ~2.5% and ~5%, indicating instability under weak amplification due to unpredictable decoder loads and CPU scheduling. Mid-range gains showed reduced prediction error, but despite these improvements, robustness remained limited. Model 2 introduced dropout, L2 weight decay, and batch normalization. Its prediction performance consistently maintained values <1.5%, outperforming Model 1. Model 3, the hybrid pipeline, delivered the best results with the maximum error ~0.5%. In the critical low-gain region, where other models degraded, errors remained ~0.1%. These results demonstrate that the hybrid model captures nonlinear scaling and discrete shifts in energy use more effectively than the other DNN models.

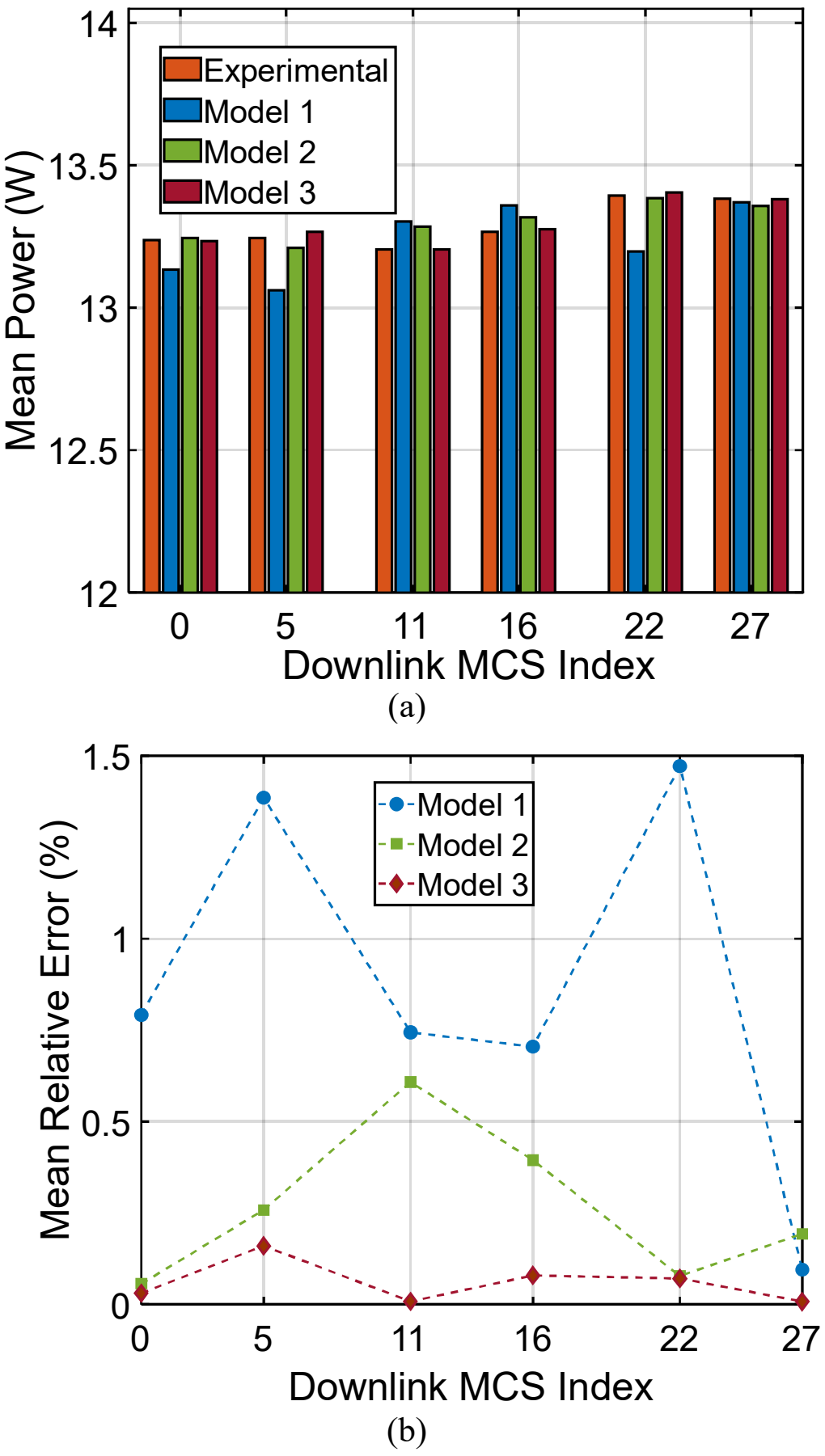

Fig. 4. Variation in (a) power consumption, and (b) relative error with increase in downlink modulation and coding scheme (MCS) index for different predictive models.

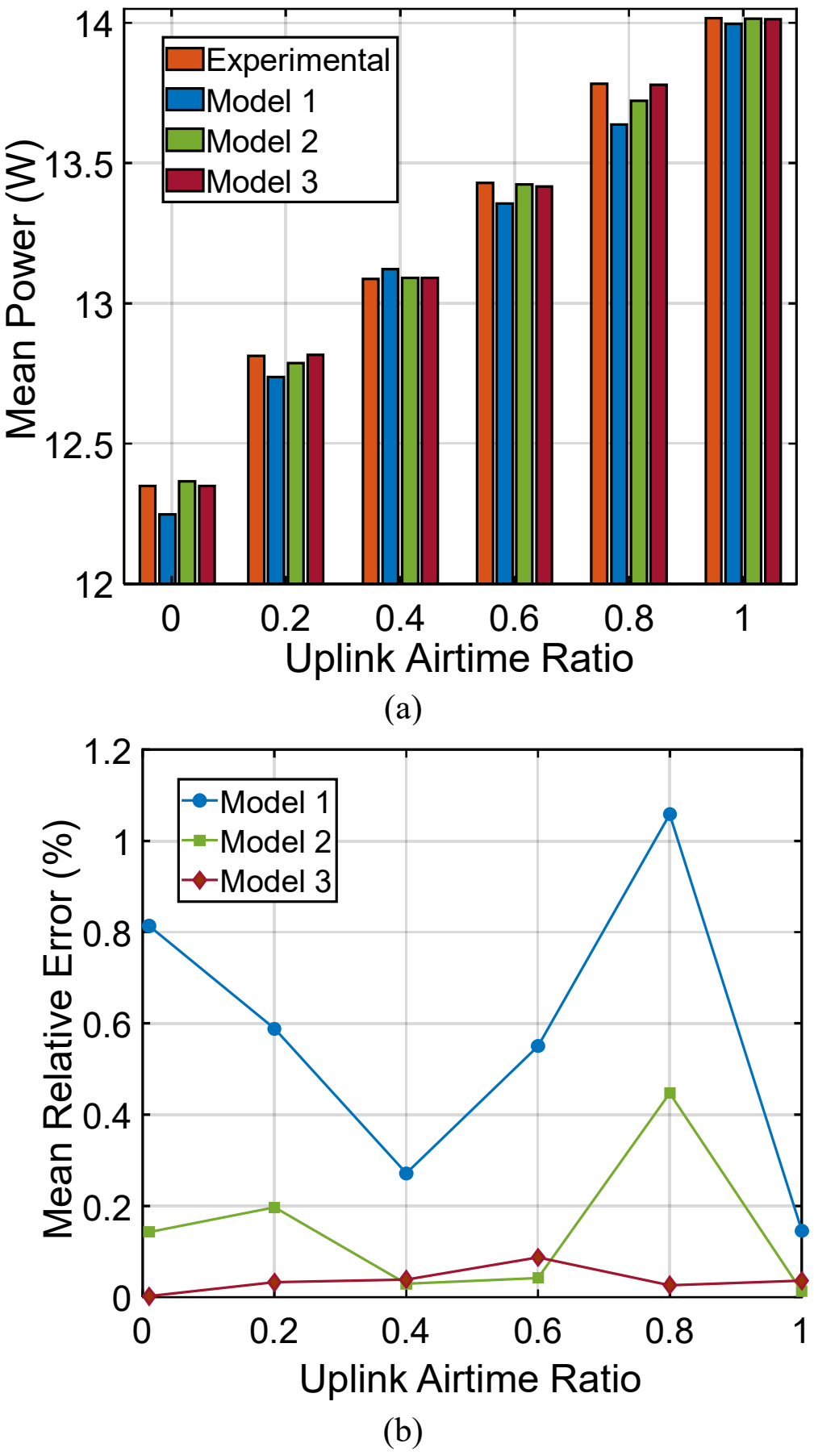

Fig. 5. Variation in (a) power consumption, and (b) relative error with increase in uplink airtime ratio for different predictive models.

## B. Modulation and Coding Scheme (MCS)

MCS determines how data is encoded and transmitted over the network. MCS indices depend largely on the channel state information and determine the efficiency of data transmission and directly influence energy consumption. Higher MCS indices increase CPU processing demands because higher modulation schemes (e.g., 64-QAM, 256-QAM) require more complex encoding/decoding. Lower MCS indices reduce CPU load but transmit less data, reducing transmission efficiency [20].

We obtain the power consumption predicted by the three models and the corresponding MRE for six different MCS settings {0, 5, 9, 14, 18, 23} in the uplink. As shown in Fig. 3, Model 3 performed best overall, with <0.1% error for most indices. In the mid-range, errors stayed negligible, rising modestly to ~0.1% at higher indices. In Fig. 4, we depict the predictive performance for downlink MC settings {0, 5, 11, 16, 22, 27}. Model 1 achieved errors up to ~1.5% which reduced at higher indices. Model 2 consistently reduced errors to <0.6%, improving reliability across all indices. Model 3 delivered the most stable results, with errors <0.25% across all levels. Its hybrid approach effectively modeled both radio conditions and processing demand.

## C. Airtime

Airtime refers to how much of the network's time-frequency resources are allocated to a device for sending (uplink) or receiving (downlink) data at the base station. It is measured in terms of the ratio or percentage of physical resource blocks allocated to a device [14]. For both, uplink and downlink, we obtain the predicted power consumption and MRE values for different airtime ratios in the range [0, 1]. In Fig. 5, we analyze the prediction performance with respect to uplink airtime ratios. Model 1 achieved errors between ~0.2% and ~1.1%. Model 2 lowered errors to <0.5% across all airtime levels. Model 3 outperformed both, consistently keeping errors <0.1%. Unlike Models 1 and 2, it avoided spikes under high-airtime conditions, maintaining stability across all load scenarios. In Fig. 6, the downlink airtime ratio is varied. Model 1 produced acceptable errors (<1%) but reached 1.5% under near-idle states (zero airtime). Model 2 improved accuracy, reducing error from ~0.5% to ~0.2% at higher values. Model 3 consistently achieved <0.1% across all levels, adapting well to low-utilization conditions.

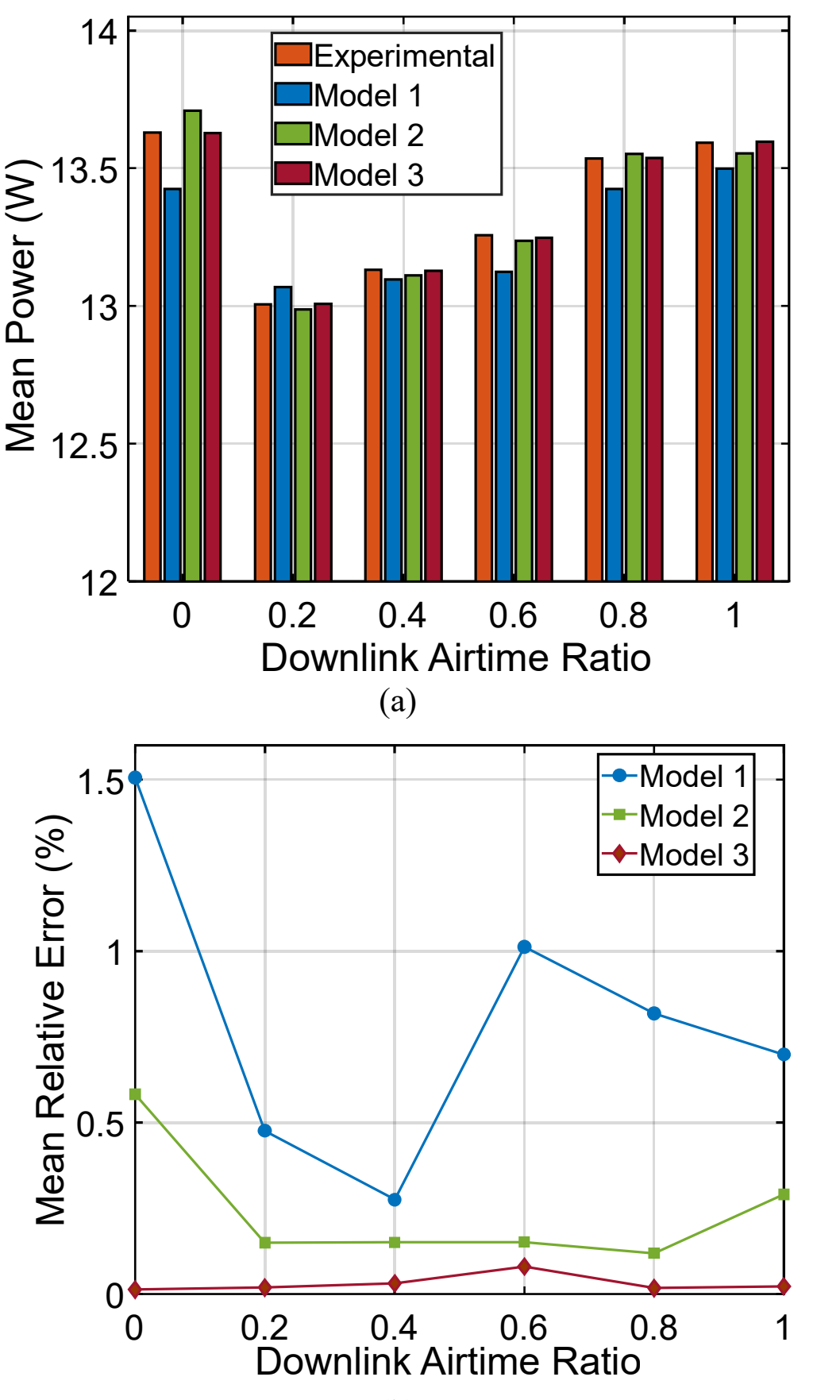

Fig. 6. Variation in (a) power consumption, and (b) relative error with increase in downlink airtime ratio for different predictive models.

## V. Conclusion

We study DNN-based ML architectures to predict power consumption in virtualized O-RAN systems. We showed that the hybrid DNN–XGBoost model achieved the highest accuracy, with mean relative error consistently below 0.5% across diverse configurations. By evaluating models on grouped radio parameters rather than random splits, we uncovered system-level insights for different system parameters like transmission gain, MCS index and airtime ratio. The proposed predictive models can be deployed to enable energy-aware operations such as adaptive transmission power control and selective radio chain deactivation. Looking ahead, we identify two promising research directions: (i) incorporating temporal dynamics using sequential learning to capture traffic-dependent energy transitions, and (ii) applying transfer learning from large-scale pretrained models to enhance prediction accuracy when labeled RAN data is limited. Together, these contributions establish a practical baseline for machine learning–based energy prediction in O-RAN and provide a foundation for future work in energy-efficient 5G-Advanced and 6G networks.

## Acknowledgements

This work was supported in part by Research Ireland grants 18/CRT/6222 and 13/RC/2077/P2; and by the EU projects 6G-XCEL, ECO-eNET and ENGCoN. The projects 6G-XCEL and ECO-eNET have received funding from the SNS JU under grant nos. 101139194 and 101139133, respectively, and ENGCoN has received funding from the MSCA-PF scheme under grant no. 101155602.